*Research Article*

# A Deep Evolutionary Approach to Bioinspired Classifier Optimisation for Brain-Machine Interaction

**Jordan J. Bird**, **Diego R. Faria, Luis J. Manso,**
**Anikó Ekárt, and Christopher D. Buckingham**

*School of Engineering and Applied Science, Aston University, Birmingham B4 7ET, UK*

Correspondence should be addressed to Jordan J. Bird; birdj1@aston.ac.uk

This study suggests a new approach to EEG data classification by exploring the idea of using evolutionary computation to both select useful discriminative EEG features and optimise the topology of Artificial Neural Networks. An evolutionary algorithm is applied to select the most informative features from an initial set of 2550 EEG statistical features. Optimisation of a Multilayer Perceptron (MLP) is performed with an evolutionary approach before classification to estimate the best hyperparameters of the network. Deep learning and tuning with Long Short-Term Memory (LSTM) are also explored, and Adaptive Boosting of the two types of models is tested for each problem. Three experiments are provided for comparison using different classifiers: one for attention state classification, one for emotional sentiment classification, and a third experiment in which the goal is to guess the number a subject is thinking of. The obtained results show that an Adaptive Boosted LSTM can achieve an accuracy of 84.44%, 97.06%, and 9.94% on the attentional, emotional, and number datasets, respectively. An evolutionary-optimised MLP achieves results close to the Adaptive Boosted LSTM for the two first experiments and significantly higher for the number-guessing experiment with an Adaptive Boosted DEvo MLP reaching 31.35%, while being significantly quicker to train and classify. In particular, the accuracy of the nonboosted DEvo MLP was of 79.81%, 96.11%, and 27.07% in the same benchmarks. Two datasets for the experiments were gathered using a Muse EEG headband with four electrodes corresponding to TP9, AF7, AF8, and TP10 locations of the international EEG placement standard. The EEG MindBigData digits dataset was gathered from the TP9, FP1, FP2, and TP10 locations.

## 1. Introduction

Bioinspired algorithms have been extensively used as robust and efficient optimisation methods. Despite the fact that they have been criticised for being computationally expensive, they have also been proven useful to solve complex optimisation problems. With the increasing availability of computing resources, bioinspired algorithms are growing in popularity due to their effectiveness at optimising complex problem solutions. Scientific studies of natural optimisation from many generations past, such as Darwinian evolution, are now becoming viable inspiration for solving real-world problems.

This increasing resource availability is also allowing for more complex computing in applications such as Internet of Things (IoT), Human-Robot Interaction (HRI), and human-computer interaction (HCI), providing more degrees of both control and interaction to the user. One of these degrees of control is the source of all others, the human brain, and it can be observed using electroencephalography. At its beginning EEG was an invasive and uncomfortable method, but with the introduction of dry, commercial electrodes, EEG is now fully accessible even outside of laboratory setups.

It has been noted that a large challenge in brain-machine interaction is inferring the attentional and emotional states from particular patterns and behaviours of electrical brain activity. Large amounts of data are needed to be acquired from EEG, since the signals are complex, nonlinear, and nonstationary. To generate discriminative features to describe a wave requires the statistical analysis of time window intervals. This study focuses on bringing together previous related research and improving the state-of-the-art with a Deep Evolutionary (DEvo) approach when optimising bioinspired classifiers. The application of this study allows for a whole bioinspired and optimised approach for mental attention classification and emotional state classification and to guess the number in which a subject thinks of. These states can then be taken forward as states of control in, for example, human-robot interaction.

In addition to the experimental results, the contributions of the work presented in this paper are as follows:

(i) An effective framework for classification of complex signals (brainwave data) through processes of evolutionary optimisation and bioinspired classification.

(ii) A new evolutionary approach to hyperheuristic bioinspired classifiers to prevent convergence on local minima found in the EEG feature space.

(iii) To gain close to identical accuracies, and in one case exceeding them, with resource-intensive deep learning through the optimised processes found in nature.

The remainder of this article proceeds as follows: Section 2 provides an exploration of the state-of-the-art works related to this study, briefly introducing the most relevant concepts applied into the DEvo approach to machine learning with electroencephalographic data. Section 3 describes the methods used to perform the experiments performed. The results of the experiments, including graphical representations of results and discussion of implications, are presented in Section 4. Section 5 details the conclusions extracted from the experiments and the suggested future work.

## 2. Background

*2.1. Electroencephalography and Machine Learning with EEG.* Electroencephalography, or EEG, is the measurement and recording of electrical activity produced by the brain [4]. The collection of EEG data is carried out through the use of applied electrodes, which read minute electrophysiological currents produced by the brain due to nervous oscillation [5, 6]. The most invasive form of EEG is subdural [7] in which electrodes are placed directly on the brain itself. Far less invasive techniques require electrodes to be placed around the cranium, of which the disadvantage is that signals are being read through the thick bone of the skull [8]. Raw electrical data is measured in microVolts (uV), which over time produce wave patterns.

Lövheim's 2012 study produced a new three-dimensional way of graphically representing human emotion in terms of categories and hormone levels [9]. This graphical representation can be seen in Figure 1, with exposition of emotional categories found in Table 1. Each vertex of the cube represents a centroid of an emotional category. It is worth noting that categories are not completely concrete, and that emotions are experienced in gradient, as well as overlapping between categories [10]. It is this chemical composition that causes certain nervous oscillation and thus electrical brainwave activity [11]. Thus, the brainwave activity can be used as data to estimate human emotions.

The Muse headband is a commercially available EEG recording device with four electrodes placed on the TP9,

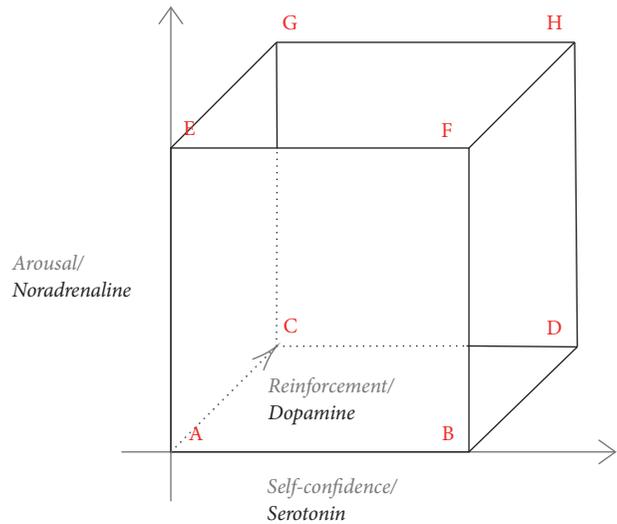

Figure 1: Lövheim's cube: mapping levels of noradrenaline, dopamine, and serotonin to human emotion.

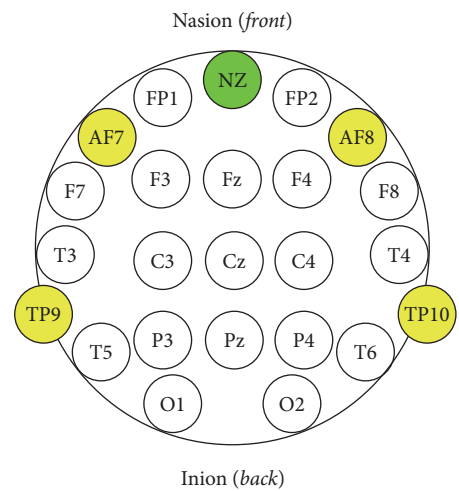

Figure 2: EEG sensors TP9, AF7, AF8, and TP10 of the Muse headband on the international standard EEG placement system [1].

AF7, AF8, and TP10 positions based on the international EEG placement system [1]. These can be seen in Figure 2. Because the signals are quite weak in nature, signal noise is a major issue due to it effectively masking the useful information [12]. The EEG headband employs various artefact separation techniques to best retain the brainwave data and discard unwanted noise [13]. Previously, the headband has been used along with machine learning techniques to measure different levels of user enjoyment, treating it as a gradient much like in sentiment analysis projects, researchers successfully managed to measure different levels of a user's enjoyment [14, 15] while playing mobile phone games. Muse headbands are also often used in neuroscience research projects due to their low-cost and ease of deployment (since they are a consumer product), as well as its effectiveness in terms of classification and accuracy [16]. In this experiment, binary classification of two physical tasks achieved 95% accuracy using Bayesian probability methods.

Previous work with the Muse headband used classical and ensemble machine learning techniques to accurately classify both mental [17] and emotional [3] states based on datasets generated by statistical extraction. The application of statistical extraction as a form of data preprocessing is useful across many platforms, e.g., for semantic place recognition in human-robot interaction [18, 19]. Machine learning techniques with inputs being that of statistical features of the wave are commonly used to classify mental states [20, 21] for brain-machine interaction, where states are used as dimensions of user input. Probabilistic methods such as Deep Belief Networks, Support Vector Machines, and various types of neural network have been found to experience varying levels of success in emotional state classification, particularly in binary classification [22].

EEG brainwave data classification is a contemporary focus in the medical fields; abnormalities in brainwave activity have been successfully classified as those leading to a stroke using a Random Forest classification method [23]. In addition to the detection of a stroke, researchers also found that monitoring classified brain activity aided successfully with rehabilitation of motor functions after stroke when coupled with human-robot interaction [24]. Brainwave classification has also been very successful in the preemptive detection of epileptic seizures in both adults and newborn infants [25, 26]. The classification of minute parts of the sleep-wake cycle is also a focus of medical researchers in terms of EEG data mining. Low resolution, three-state (*awake, sleep,* and *REM sleep*) EEG data was classified with Bayesian methods to a high accuracy of 92-97% in both humans and rats using identical models [27], both showing the ease of classification of these states as well as the cross-domain application between human and rat brains. Random Forest classification of an extracted set of statistical EEG attributes could classify sleeping patterns with higher resolution than that of the previous study at around 82% accuracy [28]. It is worth noting that for a real-time averaging technique (prediction of a time series of, for example, every 1 second), only majority classification accuracies at >50% would be required, though the time series could be trusted at shorter lengths with better results from the model.

Immune Clonal Algorithm, or ICA, has been suggested as a promising method for EEG brainwave feature extraction through the generation of mathematical temporal wave descriptors [29]. This approach found success in classification of epileptic brain activity through generated features as inputs to Naive Bayes, Support Vector Machine, K-Nearest Neighbours, and Linear Discriminant Analysis classifiers.

Autonomous classification through affective computing in human-machine interaction is a very contemporary area of research due to the increasing amounts of computational resources available, including, but not limited to, facial expression recognition [30], Sentiment Analysis [31], human activity recognition [32, 33], and human behaviour recognition [34, 35]. In terms of social human-machine interaction, a Long Short-Term Memory network was found to be extremely useful in user text analysis to derive an affective sentiment based on negative and positive polarities [36] and was used in the application of a chatbot.

Table 1: Exposition of emotional categories of Figure 1.

| Emotion Category | Emotion |
| --- | --- |
| A | Shame |
| | Humiliation |
| B | Contempt |
| | Disgust |
| C | Fear |
| | Terror |
| D | Enjoyment |
| | Joy |
| E | Distress |
| | Anguish |
| F | Surprise |
| G | Anger |
| | Rage |
| H | Interest |
| | Excitement |

*2.2. Evolutionary Algorithms.* An evolutionary algorithm will search a problem space inspired by the natural process of Darwinian evolution [37]. Solutions are treated as living organisms that, as a population, will produce more offspring that can survive. Where each solution has a measurable *fitness*, a *survival of the fittest* will occur, causing the weaker solutions to be killed off and allowing for the stronger to survive [38]. The evolutionary search in its simplest form will follow this process:

(1) Create an initial random population solution
(2) Simulate the following until termination occurs:
   (a) Using a chosen method, select parent(s) for use in generating offspring(s)
   (b) Evaluate the offspring's fitness
   (c) Consider the whole population, and kill off the weakest members

The aforementioned algorithm is often used to decide on network parameters [39] since there is *"no free lunch"* [40] when it comes to certain types of optimisation problems. In particular, it has been demonstrated that the problem of searching for the optimal parameters for a neural network cannot be solved in polynomial time [41].

*2.3. Multilayer Perceptron.* A Multilayer Perceptron is a type of Artificial Neural Network (ANN) that can be used as a universal function approximator and classifier. It computes a number of inputs through a series of layers of neurons, finally outputting a prediction of class or real value. More than one hidden layer forms a *deep neural network*. Output nodes are the classes used for classification with a softmax (single) choice, or, if there is just one a regression output (e.g., stock price prediction in GBP).

Learning is performed for a defined time measured in epochs and follows the process of *backpropagation* [42]. Backpropagation is a case of automatic differentiation in which errors in classification or regression (when comparing outputs of a network to ground truths) are passed backwards from the final layer, to derive a gradient which is then used to calculate neuron weights within the network, dictating their activation. That is, a gradient descent optimisation algorithm is employed for the calculation of neuron weights by computing the gradient of the loss function (error rate). After learning, a more optimal neural network is generated which is employed as a function to best map inputs to outputs or attributes to class.

The process of weight refinement for the set training time is given as follows:

(1) Generate the structure of the network based on input nodes, defined hidden layers, and required outputs.

(2) Initialise all of the node weights randomly.

(3) Pass the inputs through the network and generate predictions as well as cost (errors).

(4) Compute gradients.

(5) Backpropagate errors and adjust neuron weights.

Errors can be calculated in numerous ways, e.g., distance in Euclidean or non-Euclidean space for regression. In classification problems, entropy is often used, that is, the level of randomness or predictability for the classification of a set:

$$E(s) = -\sum_j p_j \times \log(p_j) \quad (1)$$

Comparing the difference of two measurements of entropy (two models) gives the information gain (*relative entropy*). This is the value of the Kullback-Leibler (KL) divergence when a univariate probability distribution of a given attribute is compared to another [43]. The calculation with the entropy algorithm in mind is thus simply given as

$$InfoGain(T, a) = H(T) - H(T \mid a) \quad (2)$$

A positive information gain denotes a lower error rate and thus a better model, i.e., a more improved matrix of network weights.

Denser is a related novel method of evolutionary optimisation of an MLP [44]. Whereas this study focuses on the search space of layer structure within fully connected neural networks, Denser also considers the type of layer. This increase of parameters to optimise grows the search space massively and is a very computationally intensive algorithm, which achieves very high results. Benchmarked is an impressive result of 93.29 on the CIFAR-10 image recognition dataset. EvoDeep [45] is a similar approach focusing on deep neural networks with varying layers; researchers found Roulette Selection (random) to be the best for selecting two parents for offspring, and thus such selection was chosen for this study's evolutionary search. A method of *"Extreme Learning Machines"* was proposed for the optimisation of deep learning processes and was extended to also perform feature extraction within the topological layers of the model [46].

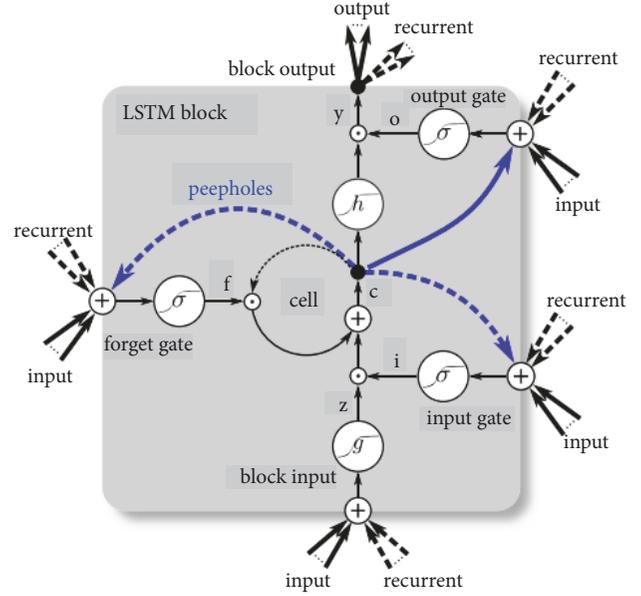

Figure 3: Diagram of a standard block within a Long Short-Term Memory network [2].

### 2.4. Long Short-Term Memory.

Long Short-Term Memory (LSTM) is a form of Artificial Neural Network in which multiple Recurrent Neural Networks (RNN) will predict based on state and previous states. As seen in Figure 3, the data structure of a neuron within a layer is an "LSTM Block". The general idea is as follows.

#### 2.4.1. Forget Gate.
The forget gate will decide on which information to store and which to delete or *"forget"*:

$$f_t = \sigma\left(W_f \cdot [h_{t=1}, \; x_t + b_f]\right), \quad (3)$$

where t is the current timestep, Wf is the matrix of weights, h is the previous output (*t-1*), xt is the batch of inputs as a single vector, and finally bf is an applied bias.

#### 2.4.2. Data Storage and Hidden State.
After deciding which information to forget, the unit must also decide which information to remember. In terms of a cell input *i*, Ct is a vector of new values generated.

$$o_t = \sigma\left(W_i \cdot [h_{t=1}, \; x_t + b_i]\right), \quad (4)$$

$$\widetilde{C}_t = \tanh\left(W_c \cdot [h_{t=1}, \; x_t + b_c]\right). \quad (5)$$

Using the calculated variables in the previous operations, the unit will follow a convolutional operation to update parameters:

$$C_t = f_t * C_{t-1} + i_t * \widetilde{C}_t. \quad (6)$$

#### 2.4.3. Output.
In the final step, the unit will produce an output at output gate Ot after the other operations are

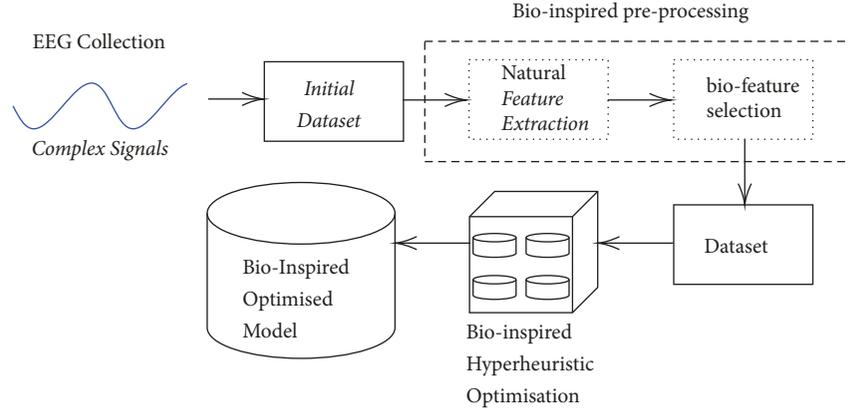

FIGURE 4: A graphical representation of the Deep Evolutionary (DEvo) approach to complex signal classification. An evolutionary algorithm simulation selects a set of natural features before a similar approach is used, then this feature set becomes the input to optimise a bioinspired classifier.

complete, and the hidden state of the node is updated:

$$o_t = \sigma\left(W_o \cdot [h_{t=1}, \ x_t + b_o\right), \quad (7)$$

$$h_t = o_t * \tanh\left(C_t\right). \quad (8)$$

Due to the observed consideration of time sequences, i.e., previously seen data, it is often found that time dependent data (waves; logical sequences) are very effectively classified thanks to the addition of unit memory. LSTMs are particularly powerful when dealing with speech recognition [47] and brainwave classification [48] due to their temporal nature.

*2.5. Adaptive Boosting.* Adaptive Boosting (AdaBoost) is an algorithm which will create multiple unique instances of a certain model to attempt to mitigate situations in which selected parameters are less effective than others at a certain time [49]. The models will combine their weighted predictions after training on a random data subset to improve the previous iterations. The fusion of models is given as

$$F_T(x) = \sum_{t=1}^{T} f_t(x), \quad (9)$$

where $F$ is the set of classifiers and $x$ is the data object being considered [50].

## 3. Method

Building on top of previous works which have succeeded using bioinspired classifiers for prediction of biological processes, this work suggests a completely bioinspired process. It includes biological inspiration into every step of the process rather than just the classification stage. The system as a whole therefore has the following stages:

(1) Generation of an initial dataset of biological data, EEG signals in particular (*collection*).

(2) Selection of attributes via biologically inspired computing (*attribute selection*).

(3) Optimisation of a neural network via biologically inspired computing (*hyperheuristics*).

(4) Use of an optimised neural network for the classification of the data (*classification*).

The steps allow for evolutionary optimisation of data preprocessing as well as using a similar approach for deep neural networks which also evolve. This leads to the *Deep Evolutionary* or *DEvo* approach. A graphical representation of the above steps can be seen in Figure 4. Nature is observed to be close to optimal in both procedure and resources; the goal of this process therefore is to best retain high accuracies of complex models, but to reduce the processing time required to execute them.

The rest of this section serves to give details to the steps of the DEvo approach seen in Figure 4.

*3.1. Data Acquisition.* As previously mentioned, the paper at hand provides three experiments dealing with the classification of the attentional, emotional state, and *"thinking of"* state of subjects. For the first two sets of experiments, two datasets were acquired from previous studies [3, 17]. The first dataset (mental state) distinguishes three different states related to how focused the subject is: relaxed, concentrative, or neutral (https://www.kaggle.com/birdy654/eeg-brainwave-dataset-mental-state). This data was recorded for three minutes, per state, per person of the subject group. The subject group was made up of two adult males and two adult females aged 22 ± 2. The second dataset (emotional state) was based on whether a person was feeling positive, neutral, or negative emotions (https://www.kaggle.com/birdy654/eeg-brainwave-dataset-feeling-emotions). Six minutes for each state were recorded from two adults, 1 male and 1 female aged 21 ± 1 producing a total of 36 minutes of brainwave activity data. The experimental setup of the Muse headset being used to gather data from the TP9, AF7, AF8, and TP10 extra-cranial electrodes during a previous study [3] can be seen in Figure 5. An example of the raw data retrieved from the headband can be seen in Figure 6. Additionally, observations of the range of subjects for the two aforementioned datasets were made; educational level was relatively high within the subjects, two were PhD Students, one Master's Student, and one with a BSc degree, all from STEM fields. All subjects were in fine health, both physical and mental. All subjects were from the United Kingdom, three were from the West Midlands whereas one was from Essex. All of the subjects volunteered to take part in this study.

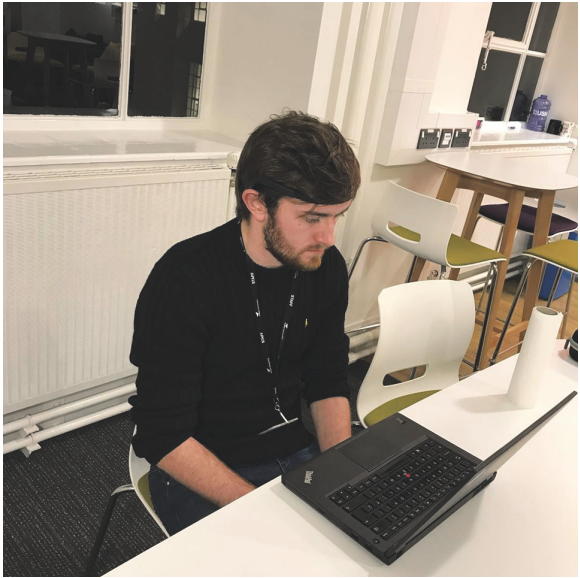

Figure 5: A subject having their EEG brainwave data recorded while being exposed to a stimulus with an emotional valence [3].

The two mental state datasets are a constant work in progress in order to become representative of a whole human population rather than those described in this section, the data as-is provides a preliminary point of testing and a proof of concept of the DEvo approach to bioinspired classifier optimisation, and this would be an ongoing process if subject diversity has a noticeable impact, since the global demographic often changes.

For the third experiment, the "MindBigData" dataset was acquired and processed (http://www.mindbigdata.com/opendb/). This publicly available data is an extremely large dataset gathered over the course of two years from one subject in which the subject was asked to think of a digit between and including 0 to 9 for two seconds. This gives a ten class problem. Due to the massive size of the dataset and computational resources available, 15 experiments for each class were extracted randomly, giving a uniform extraction of 30 seconds per digit class and therefore 300 seconds of EEG brainwave data. It must be critically noted that a machine learning model would be classifying this single subject's brainwaves, and in conjecture, transfer learning is likely impossible. Future work should concern the gathering of similar data but from a range of subjects. The MindBigData dataset used a slightly older version of the Muse headband, corresponding to two slightly different yet still frontal lobe sensors, collecting data from the TP9, FP1, FP2, and TP10 electrode locations.

*3.2. Full Set of Features (Preselection).* As described previously, feature extraction is based on previous research into effective statistical attributes of EEG brainwave data [17]. This section describes the reasoning behind the necessity of performing statistical extraction, as well as the method to perform the process.

The EEG sensor used for the experiments, the Muse headband, communicates with the computer using Bluetooth Low energy (BLE). The use of this protocol improves the autonomy of the sensor at the expense of a nonuniform sampling rate. The first step applied to normalise the dataset is using a Fourier-based method to resample the data to a fixed frequency of 200Hz.

Brainwave data is nonlinear and nonstationary in nature, and thus single values are not indicative of class. That is, mental classification is based on the temporal nature of the wave, and not the values specifically. For example, a simplified concentrative and relaxed wave can be visually recognised due to the fact that wavelengths of concentrative mental state class data are far shorter, and yet, a value measured at any one point might be equal for the two states (i.e., $x$ microVolts). Additionally, the detection of the natures that dictate alpha, beta, theta, delta, and gamma waves also requires analysis over time. It is for these reasons that temporal statistical extraction is performed. For temporal statistical extraction, sliding time windows of total length 1s are considered, with an overlap of 0.5 seconds. That is, windows run from $[0s - 1s), [1.5s - 2.5s), [2s - 3s), [2.5s - 3s)$, continuing until the experiment ends.

The remainder of this subsection describes the different statistical features types which are included in the initial dataset:

(i) A set of values of signals within a sequence of temporal windows $x_1, x_2, x_3 \cdots x_n$ are considered and mean values are computed:

$$\mu \frac{1}{N} \sum_N^i x_i. \quad (10)$$

(ii) The standard deviation of values is recorded:

$$\sigma = \sqrt{\frac{1}{N} \sum_N^i (x_i - \mu)^2}. \quad (11)$$

(iii) Asymmetry and peakedness of waves are statistically represented by the skewness and kurtosis via the statistical moments of the third and fourth order. Skewness:

$$y = \frac{\mu^k}{\sigma^k} \quad (12)$$

and kurtosis:

$$\mu^k = \frac{1}{N} \sum_N^i (x_i - \mu)^k \quad (13)$$

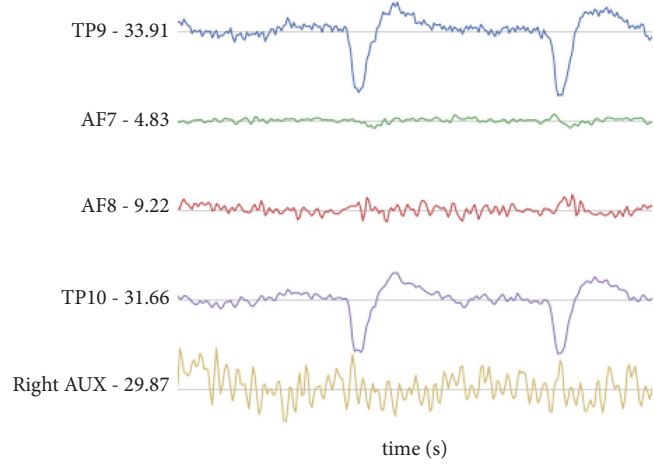

FIGURE 6: An example of a raw EEG data stream from the Muse EEG headband. The Y-axis represents measured brainwave activity in microVolts (mV) and the X-axis is the time at which the data was recorded.

are taken where k=3rd and k=4th moment about the mean.

(iv) Max value within each particular time window $\{max_1, max_2, \ldots, max_n\}$.

(v) Minimum value within each particular time window $\{min_1, min_2, \ldots, min_n\}$.

(vi) Derivatives of the minimum and maximum values by dividing the time window in half, and measuring the values from either half of the window.

(vii) Performing the min and max derivatives a second time on the presplit window, resulting in the derivatives of every 0.25s time window.

(viii) For every min, max, and mean value of the four 0.25s time windows, the Euclidean distance between them is measured. For example, the maximum value of time window one of four has its 1D Euclidean distance measured between it and max values of windows two, three, and four of four.

(ix) From the 150 features generated from quarter-second min, max, and mean derivatives, the last six features are ignored and thus a 12x12 (144) feature matrix can be generated. Using the Logarithmic Covariance matrix model [51], a log-cov vector and thus statistical features can be generated for the data as such

$$lcM = U\left(\log m\left(\text{cov}\left(M\right)\right)\right). \quad (14)$$

U returns the upper triangular features of the resultant vector and the covariance matrix (cov(M)) is

$$\text{cov}\left(M\right) = \text{cov}_{ij} = \frac{1}{N\sum_{N}^{k}\left(x_{ik} - \mu_i\right)\left(x_{kj} - \mu_j\right)}. \quad (15)$$

(x) For each full 1s time window, the Shannon Entropy is measured and considered as a statistical feature:

$$h = -\sum_{j} S_j \times \log\left(S_j\right). \quad (16)$$

The complexity of the data is summed up as such, where h is the statistical feature and S relates to each signal within the time window after normalisation of values.

(xi) For each 0.5s time window, the log-energy entropy is measured as

$$\log e = \sum_{i} \log\left(S_i^2\right) + \sum_{j} \log\left(S_i^2\right), \quad (17)$$

where $i$ is the first time window $n$ to $n+0.5$ and $j$ is the second time window $n+0.5$ to $n+1$.

(xii) Analysis of a spectrum is performed by an algorithm to perform Fast Fourier Transform (FFT) [52] of every recorded time window, derived as follows:

$$X_k = \sum_{n=0}^{N-1} S_n^t e^{-i2\pi k(n/N)}, \quad k = 0, \ldots, N-1. \quad (18)$$

The above statistical features are used to represent the waves. With these features considered for each electrode and time window (including those formed by overlaps), this produces a total of 2147 scalars per measure. The resulting number of features is too large to be used in real time (i.e., it would be computationally intensive) and would not yield good classification results because of the large dimensionality. Attribute selection is therefore performed to overcome this limitations and, additionally, make the train process significantly faster.

3.3. Evolutionary Optimisation and Machine Learning. The evolutionary optimisation process as detailed previously was applied when selecting discriminative attributes from the full dataset for more optimised classification. An initial population of 20 attribute subsets were generated and simulated for 20 generations with tournament breeding selection [53]. Evolutionary optimisation was also applied to explore the n-dimensional MLP topological search space, where n is the number of hidden layers, with the goal of searching for the best accuracy (fitness metric). With the selected attributes forming the new dataset to be used in experiments, two models were generated: an LSTM and an MLP.

Before finalising the LSTM model, various hyperparameters are explored, specifically the topology of the network. This was performed manually since evolutionary optimisation of LSTM topology would have been extremely computationally expensive. More than one hidden layer often returned worse results during manual exploration and thus one hidden layer was decided upon. LSTM units within this layer would be tested from 25 to 125 at steps of 25 units. Using a vector of the time sequence statistical data as an input in batches of 50 data points, an LSTM was trained for 50 epochs to predict class for each number of units on a layer, and thus a manually optimised topology was derived.

A Multilayer Perceptron was first fine-tuned via an evolutionary algorithm [39] with the number of neurons and layers as population solutions, with classification accuracy as a fitness. A maximum of three hidden layers and up to 100 neurons per layer were implemented into the simulation. Using 10-fold cross validation, the MLP had the following parameters manually set:

(i) 500-epoch training time
(ii) Learning rate of 0.3
(iii) Momentum of 0.2
(iv) No decay

Finally, the two models were attemptedly boosted using the AdaBoost algorithm in an effort to mitigate both the ill-effects of manually optimising the LSTM topology as well as fine-tune the models overall.

## 4. Results and Discussion

*4.1. Evolutionary Attribute Selection.* An evolutionary search within the 2550 dimensions of the datasets was executed for 20 generations and a population of 20. For mental state, the algorithm selected 99 attributes, whereas for the emotional state, the algorithm selected a far greater 500 attributes for the optimised dataset. This suggests that emotional state has far more useful statistical attributes for classification whereas mental state requires approx. 80% fewer. The MindBigData EEG problem set, incomparable due to the previous due to its larger range of classes, had 40 attributes selected by the algorithm. This can be seen in Table 2.

The evolutionary search considered the information gain (Kullback-Leibler Divergence) of the attributes and thus their classification ability as a fitness metric, i.e., where a higher information gain represents a more effective and less entropic a model when such attributes are considered as input parameters. The search selected large datasets, between sizes 40 for the MBD dataset, to the 500 selected for the emotional state dataset. Though too numerous to detail the whole process (all datasets are available freely online for full recreation of experiments), observations were as follows:

(i) For the mental state dataset, 99 attributes were selected; the highest was the entropy of the TP9 electrode within the first sliding window at an IG of 1.225. This was followed secondly with the eigenvalue of the same electrode, showing that the TP9 placement is a good indicator for concentrative states. It must be noted that these values may possibly correlate with the Sternocleidomastoid Muscle's contractional behaviours during stress and ergo the stress encountered during concentration or the lack thereof during relaxation, and thus EMG behaviours may be inadvertently classified rather than EEG.

(ii) Secondly, for the emotional state dataset, the most important attribute was observed to be the mean value of the AF7 electrode in the second overlapping time window. This gave an information gain of 1.06, closely followed by a measure of 1.05 for the first covariance matrix of the first sliding window. Minimum, mean, and covariance matrix values of electrodes all followed with IG scores from 0.98 to 0.79 until standard deviation of electrodes followed. Maximum values did not appear until the lower half of the ranked data, in which the highest max value of the second time window of the AF8 electrode had an IG of 0.66.

(iii) Finally, for the MBD dataset, few attributes were chosen. This was not due to their impressive ability, but due to the lack thereof when other attributes were observed. For example, the most effective attribute was considered the covariance matrix of the second sliding windows of the frontal lobe electrodes, FP1 and FP2, but these only has information gain values of 0.128 and 0.125 each, far lower than those observed in the other two experiments. To the lower end of the selected values, IG scores of 0.047 appear, which are considered very weak and yet still chosen by the algorithm. The MBD dataset is thus an extremely difficult dataset to classify.

Since the algorithm showed clearly a best attribute for each, a benchmark was performed using a simple One Rule Classifier (OneR). OneR will focus on the values of the best attribute and attempt to separate classes by numerical rules. In Table 3, the observations above are shown more concretely with statistical evidence. Classifying MindBigData based on the 0.128 IG attribute detailed above gains only 17.13% accuracy, whereas the far higher attributes for the other two datasets gain 49.27% and 85.27% accuracies.

The datasets generated by this algorithm are taken forward in the DEvo process, and the original datasets are thus discarded. Further experiments are performed with this data only.

*4.2. Evolutionary Optimisation of MLP.* During the algorithm's process, an issue arose with stagnation, in which the solutions would quickly converge on a local minima and an optimal solution was not found. On average, no further improvement would be made after generation 2. It can be noted that the relatively flat gradient in Figures 7 and 8 suggests that the search space's fitness matrix possibly had a much lower standard deviation and thus the area was more difficult to traverse due to the lack of noticeable peaks and troughs. The algorithm was altered to prevent genetic collapse with the addition of speciation. The changes were as follows:

(i) A solution would belong to one of three species, *A, B*, or *C*.
(ii) A solution's species label would be randomly initialised along with the population members.
(iii) During selection of *parent1's* breeding partner, only a member of *parent1's* species could be chosen.
(iv) If only one member of a species remains, it will not produce offspring.
(v) An offspring will have a small random chance to become another species (manually tuned to 5%)

Table 2: Datasets generated by evolutionary attribute selection.

| Dataset | Population | Generations | No. Chosen Attributes |
| --- | --- | --- | --- |
| Mental State | 20 | 20 | 99 |
| Emotional State | 20 | 20 | 500 |
| MindBigData | 20 | 20 | 40 |

Table 3: Accuracies when attempting to classify based on only one attribute of the highest information gain.

| Dataset | MS | ES | MBD |
| --- | --- | --- | --- |
| Benchmark Accuracy (%) | 49.27 | 85.27 | 17.13 |

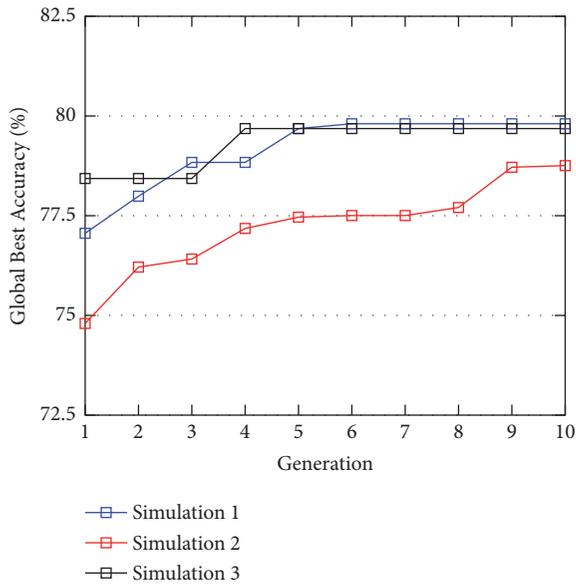

Figure 7: Three evolutionary algorithm simulations to optimise an MLP for the mental state dataset.

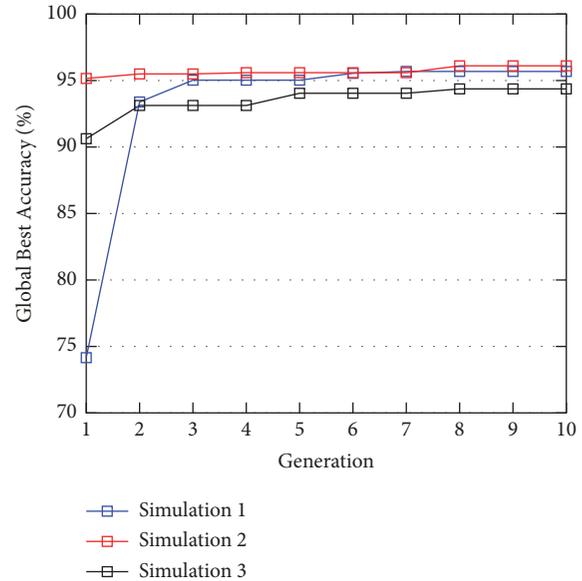

Figure 8: Three evolutionary algorithm simulations to optimise an MLP for the emotional state dataset.

The implementation of separate species in the simulation allowed for more complex, better solutions to be discovered. The increasing gradients as observed in Figures 7, 8, and 9 show that constant improvement was achieved. The evolutionary optimisation of MLP topology was set to run for a set 10 generations, tested for scientific benchmark accuracy three times due to the possibility of a single random mutation finding a good result by chance (*random search*), taking approximately ten minutes for each to execute.

This was repeated three times for purposes of scientific accuracy. Tables 4, 5, and 6 detail the accuracy values measured at each generation along with detail of the network topology. Figures 7, 8, and 9 graphically represent these experiments to detail the gradient of solution score increase.

*4.3. Manual LSTM Tuning.* Manual tuning was performed to explore the options for LSTM topology for both mental state and emotional state classification. Evolutionary optimisation was not applied due to the high resource usage of LSTM training, due to many single networks taking multiple hours to train on the 1280 CUDA cores of an NVidia GTX 1060. Results in Table 7 show that, for mental state, 100 LSTM units are somewhat most optimal, whereas 25 LSTM units were discovered to be most optimal for emotional state classification and 100 LSTM units are best for the MindBigData digit set but this result is extremely low for a uniform 10-class problem, with very little information gain. Comparison of the LSTM units to accuracy for both states can be seen in Figure 10. For each of the experiments, these arrangements of LSTM architecture will be taken forward as the selected model.

Additionally, empirical testing found that 50 epochs for training of units seemed best but further exploration is required to fine-tune this parameter. A batch size of 50 formed the input vectors of sequential statistical brainwave data for the LSTM. Gradient descent was handled by the Adaptive Moment Estimation (Adam) algorithm, with a decay value of 0.9. Weights were initialised by the commonly used XAVIER algorithm. Optimisation was performed by Stochastic Gradient Descent. Manual experiments found that a network with a depth of 1 persistently outperformed deeper networks of two or more hidden layers for this specific context; interestingly, this too is mirrored in the evolutionary optimisation algorithms for the MLP which always converged to a single layer to achieve higher fitness.

Table 4: Global best MLP solutions for mental state classification.

| Experiment | | Generation | | | | | | | | | |
|---|---|---|---|---|---|---|---|---|---|---|---|
| | | 1 | 2 | 3 | 4 | 5 | 6 | 7 | 8 | 9 | 10 |
| 1 | Layers | 1 | 1 | 1 | 1 | 1 | 1 | 1 | 1 | 1 | 1 |
| | Neurons | 17 | 13 | 26 | 26 | 28 | 27 | 27 | 27 | 27 | 27 |
| | Accuracy (%) | 77.0598 | 77.9889 | 78.8368 | 78.8368 | 79.685 | 79.8061 | 79.8061 | 79.8061 | 79.8061 | **79.8061** |
| 2 | Layers | 1 | 2 | 1 | 2 | 2 | 1 | 1 | 1 | 2 | 2 |
| | Neurons | 3 | 5, 18 | 4 | 7, 3 | 7, 9 | 8 | 8 | 18 | 9, 21 | 11, 3 |
| | Accuracy (%) | 74.7981 | 76.2116 | 76.4136 | 77.1809 | 77.4637 | 77.5040 | 77.5040 | 77.7059 | 78.7157 | 78.7561 |
| 3 | Layers | 1 | 1 | 1 | 1 | 1 | 1 | 1 | 1 | 1 | 1 |
| | Neurons | 10 | 10 | 10 | 28 | 28 | 28 | 28 | 28 | 28 | 28 |
| | Accuracy (%) | 78.4329 | 78.4329 | 78.4329 | 79.685 | 79.685 | 79.685 | 79.685 | 79.685 | 79.685 | 79.685 |

Table 5: Global best MLP solutions for emotional state classification.

| Experiment | | Generation | | | | | | | | | |
|---|---|---|---|---|---|---|---|---|---|---|---|
| | | 1 | 2 | 3 | 4 | 5 | 6 | 7 | 8 | 9 | 10 |
| 1 | Layers | 3 | 3 | 2 | 2 | 2 | 1 | 1 | 1 | 1 | 1 |
| | Neurons | 2, 8, 13 | 5, 12, 8 | 17, 17 | 21, 13 | 21, 13 | 6 | 6 | 6 | 6 | 6 |
| | Accuracy (%) | 74.1557 | 93.3865 | 95.0281 | 95.0281 | 95.0281 | 95.6848 | 95.6848 | 95.6848 | 95.6848 | 95.6848 |
| 2 | Layers | 1 | 2 | 2 | 1 | 1 | 1 | 1 | 1 | 1 | 1 |
| | Neurons | 12 | 38, 19 | 38, 19 | 8 | 8 | 8 | 8 | 15 | 15 | 15 |
| | Accuracy (%) | 95.1563 | 95.4971 | 95.4971 | 95.5909 | 95.5909 | 95.5909 | 95.5909 | 96.1069 | 96.1069 | **96.1069** |
| 3 | Layers | 3 | 2 | 2 | 2 | 2 | 2 | 2 | 2 | 2 | 2 |
| | Neurons | 7, 8, 3 | 7, 8 | 7, 8 | 7, 8 | 5, 8 | 5, 8 | 5, 8 | 9, 5 | 9, 5 | 9, 5 |
| | Accuracy (%) | 90.625 | 93.125 | 93.125 | 93.125 | 94.0431 | 94.0431 | 94.0431 | 94.3714 | 94.3714 | 94.3714 |

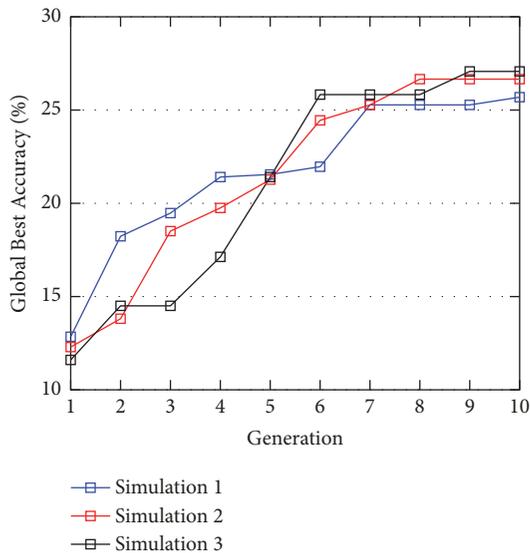

Figure 9: Three evolutionary algorithm simulations to optimise an MLP for the MindBigData dataset.

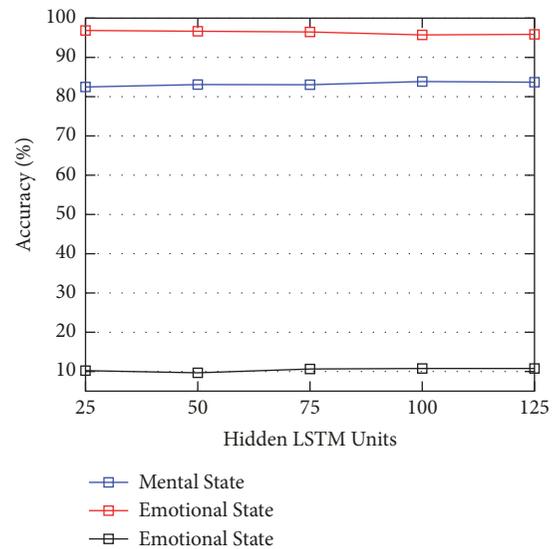

Figure 10: Manual tuning of LSTM topology for mental state (*MS*), emotional state (*ES*), and MindBigData (*MBD*) classification.

Table 6: Global best MLP solutions for MindBigData classification.

| Experiment | | Generation | | | | | | | | | |
|---|---|---|---|---|---|---|---|---|---|---|---|
| | | 1 | 2 | 3 | 4 | 5 | 6 | 7 | 8 | 9 | 10 |
| 1 | Layers | 1 | 1 | 1 | 1 | 1 | 1 | 1 | 1 | 1 | 1 |
| | Neurons | 2 | 5 | 7 | 11 | 19 | 28 | 34 | 34 | 34 | 94 |
| | Accuracy (%) | 12.8453 | 18.2320 | 19.4751 | 21.4088 | 21.5470 | 21.9613 | 25.2762 | 25.2762 | 25.2762 | 25.6906 |
| 2 | Layers | 2 | 2 | 1 | 1 | 1 | 1 | 1 | 1 | 1 | 1 |
| | Neurons | 10, 2 | 10, 9 | 9 | 10 | 25 | 27 | 34 | 87 | 87 | 87 |
| | Accuracy (%) | 12.2928 | 13.8121 | 18.5081 | 19.7514 | 21.2707 | 24.4475 | 25.2762 | 26.6575 | 26.6575 | 26.6575 |
| 3 | Layers | 2 | 2 | 2 | 1 | 1 | 1 | 1 | 1 | 1 | 1 |
| | Neurons | 3, 4 | 8, 9 | 8, 9 | 6 | 11 | 70 | 70 | 70 | 89 | 89 |
| | Accuracy (%) | 11.6022 | 14.5028 | 14.5028 | 17.1271 | 21.4088 | 25.8287 | 25.8287 | 25.8287 | 27.0718 | **27.0718** |

Table 7: Manual tuning of LSTM topology for mental state (*MS*), emotional state (*ES*), and EEG MindBigData classification.

| LSTM Units | MS (%) | ES (%) | MBD (%) |
|---|---|---|---|
| 25 | 82.47 | **96.86** | 10.22 |
| 50 | 83.08 | 96.66 | 9.67 |
| 75 | 83.04 | 96.48 | 10.64 |
| 100 | **83.84** | 95.73 | **10.77** |
| 125 | 83.68 | 95.87 | 10.36 |

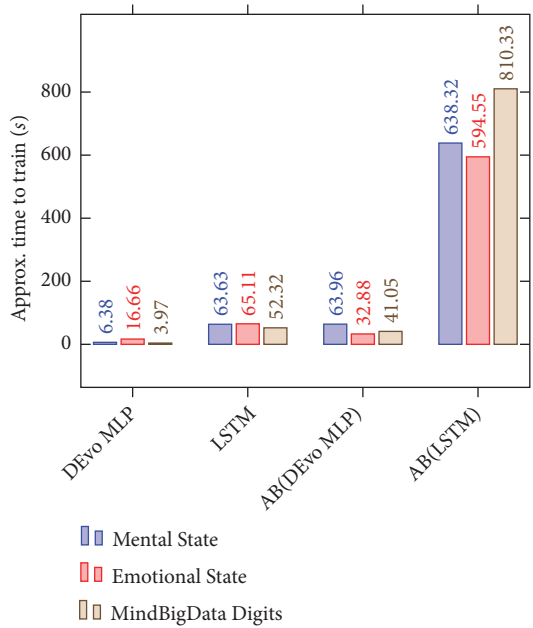

Figure 11: Graph to show the time taken to build the final models after search.

*4.4. Single and Boost Accuracy.* Figure 11 shows a comparison of approximate time taken to train the various models, note that 10-fold cross validation was performed to prevent overfitting and thus the actual time taken with this in mind is around ten times more than the displayed value. Additionally, this time was measured when training on the 1280 CUDA cores of an NVidia GTX1060 (6GB) would take considerably longer on a CPU. Although the mental state dataset had approximately five times the number of attributes, the time taken to learn on this dataset was only slightly longer than the emotional state by an average of 11% (30.26s).

Since the LSTM topology was linearly tuned in a manual process whereas the MLP was searched via an evolutionary algorithm, the processes are not scientifically comparable since the former depends on human experience and latter upon resources available. Thus, time for these processes are not given since only one is a measure of computational resource usage; it is suggested that a future study should make use of the evolutionary algorithm within the search space of LSTM topologies too, in which case they can be compared. Though, it can be inferred from Figure 11 that the search for an LSTM would take considerably longer due to the increased resources required in every experiment performed compared to the MLP. Additionally, with this in mind, a Multiobjective Optimisation (MOO) implementation of DEvo that considers both accuracy and resource usage as fitness metrics could further find more optimal models in terms of both their classification ability and optimal execution.

The overall results of the experiments can be seen firstly in Table 8 and as a graphical comparison in Figure 12. For the two three-state datasets, the most accurate model was an AdaBoosted LSTM with results of 84.44% and 97.06% accuracies for the mental state and mental emotional state datasets, respectively. The single LSTM and evolutionary-optimised MLP models come relatively close to the best result, though take far less time to train when the measured approximate values in Figure 11 are observed. On the other hand, for the MindBigData digits dataset, the best solution by far was the Adaptive Boosted DEvo MLP, and the same boosting method applied to the LSTM that previously improved them actually caused a loss in accuracy.

Manual tuning of LSTM network topology was performed due to the limited computational resources available; the success in optimisation of the MLP suggests that further improvements could be made through an automated process of evolutionary optimisation in terms of the LSTM topology. A further improvement to the DEvo system could be made by exploring the possibility of optimising the LSTM structure through an evolutionary approach. In addition, more bioinspired classification techniques should be experimented with, for example, a convolutional neural network to better imitate and improve on the classification ability of natural vision [54].

Table 8: Classification accuracy on the two optimised datasets by the DEvo MLP, LSTM, and selected boost method.

| Dataset | Accuracy (%) | | Boost Accuracy (%) | |
| --- | --- | --- | --- | --- |
| | DEvo MLP | LSTM | AB(DEvo MLP) | AB(LSTM) |
| Mental State | 79.81 | **83.84** | 79.7 | **84.44** |
| Emotional State | 96.11 | **96.86** | 96.23 | **97.06** |
| MindBigData Digits | 27.07 | 10.77 | **31.35** | 9.94 |

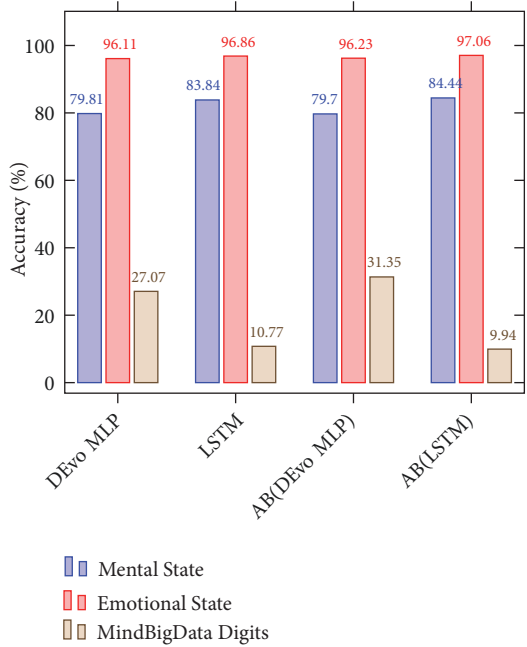

Figure 12: Final results for the experiment.

The three experiments were performed within the limitations of the Muse headband's TP9, AF7, AF8, and TP10 electrodes. Higher resolution EEG setups would allow for further exploration of the system in terms of mental data classification, e.g., for physical movement originating from the motor cortex.

## 5. Conclusion

This study suggested DEvo, a Deep Evolutionary, approach to optimise and classify complex signals using bioinspired computing methods in the whole pipeline, from feature selection to classification. For mental state and mental emotional state classification of EEG brainwaves and their mathematical features, two best models were produced:

(1) A more accurate AdaBoosted LSTM, that although it took more time and resources to train in comparison to other methods, it managed to attain accuracies of 84.44% and 97.06% for the two first datasets (attentional and emotional state classification).

(2) Secondly, a AdaBoosted Multilayer Perceptron that was optimised using a hyperheuristic evolutionary algorithm. Though its classification accuracy was slightly lower than that of the AdaBoosted LSTM (79.7% and 96.23% for the same two experiments), it took less time to train.

For the MindBigData digits dataset the most accurate model was an Adaptive Boosted version of the DEvo optimised MLP, which achieved an accuracy of 30%. For this problem, none of the LSTMs were able to achieve any meaningful or useful results, but the DEvo MLP approach saved time and also produced results that were useful. Results were impressive for application due to the high classification ability along with the reduction of resource usage; real-time training from individuals would be possible and thus provide a more accurate EEG-based product to the consumer, for example, in real-time monitoring of mental state for the grading of meditation or yoga session quality. Real-time communication would also be possible in human-computer interaction where the brain activity acts as a degree of input.

The goal of the experiment was successfully achieved, the DEvo approach has led to an optimised, resource-light model that closely matches that to an extremely resource heavy deep learning model, losing a small amount of accuracy but computing in approximately 10% of the time, except for in one case in which it far outperformed its competitor models.

The aforementioned models were trained on a set of attributes that were selected with a bioinspired evolutionary algorithm.

The success of these processes led to future work suggestions, which follow the pattern of further bioinspired optimisation applications within the field of machine learning. Future work should also consider, for better application of the process within the field of electroencephalography, a much larger collection of data from a considerably more diverse range of subjects in order to better model the classifier optimisation for the thought pattern of a global population rather than the subjects encompassed within this study.

## Data Availability

All data used in this study is freely available online; links to all datasets can be found within the data acquisition section.

## Conflicts of Interest

The authors declare that they have no conflicts of interest.

Complexity 13

## References

[1] H. H. Jasper, "The ten-twenty electrode system of the international federation," *Clinical Neurophysiology*, vol. 10, pp. 370–375, 1958.

[2] K. Greff, R. K. Srivastava, J. Koutnk, B. R. Steunebrink, and J. Schmidhuber, "LSTM: a search space odyssey," *IEEE Transactions on Neural Networks and Learning Systems*, vol. 28, no. 10, pp. 2222–2232, 2017.

[3] J. J. Bird, A. Ekart, C. D. Buckingham, and D. R. Faria, "Mental emotional sentiment classification with an eeg-based brain-machine interface," in *Proceedings of theInternational Conference on Digital Image and Signal Processing (DISP'19)*, Springer, 2019.

[4] E. Niedermeyer and F. L. da Silva, *Electroencephalography: Basic Principles, Clinical Applications, and Related Fields*, Lippincott Williams & Wilkins, 2005.

[5] A. Coenen, E. Fine, and O. Zayachkivska, "Adolf beck: a forgotten pioneer in electroencephalography," *Journal of the History of the Neurosciences*, vol. 23, no. 3, pp. 276–286, 2014.

[6] B. E. Swartz, "The advantages of digital over analog recording techniques," *Electroencephalography and Clinical Neurophysiology*, vol. 106, no. 2, pp. 113–117, 1998.

[7] A. K. Shah and S. Mittal, "Invasive electroencephalography monitoring: Indications and presurgical planning," *Annals of Indian Academy of Neurology*, vol. 17, no. 1, pp. S89–S94, 2014.

[8] B. A. Taheri, R. T. Knight, and R. L. Smith, "A dry electrode for EEG recording," *Electroencephalography and Clinical Neurophysiology*, vol. 90, no. 5, pp. 376–383, 1994.

[9] H. Lövheim, "A new three-dimensional model for emotions and monoamine neurotransmitters," *Medical Hypotheses*, vol. 78, no. 2, pp. 341–348, 2012.

[10] K. Oatley and J. M. Jenkins, *Understanding Emotions*, Blackwell Publishing, 1996.

[11] J. Gruzelier, "A theory of alpha/theta neurofeedback, creative performance enhancement, long distance functional connectivity and psychological integration," *Cognitive Processing*, vol. 10, no. 1, pp. 101–109, 2009.

[12] E.-R. Symeonidou, A. D. Nordin, W. D. Hairston, and D. P. Ferris, "Effects of cable sway, electrode surface area, and electrode mass on electroencephalography signal quality during motion," *Sensors*, vol. 18, no. 4, p. 1073, 2018.

[13] A. S. Oliveira, B. R. Schlink, W. D. Hairston, P. König, and D. P. Ferris, "Induction and separation of motion artifacts in EEG data using a mobile phantom head device," *Journal of Neural Engineering*, vol. 13, no. 3, p. 036014, 2016.

[14] M. Abujelala, A. Sharma, C. Abellanoza, and F. Makedon, "Brain-EE: Brain enjoyment evaluation using commercial EEG headband," in *Proceedings of the 9th ACM International Conference on Pervasive Technologies Related to Assistive Environments, PETRA 2016*, p. 33, ACM, Greece, July 2016.

[15] A. Plotnikov, N. Stakheika, A. De Gloria et al., "Exploiting real-time EEG analysis for assessing flow in games," in *Proceedings of the 12th IEEE International Conference on Advanced Learning Technologies, ICALT 2012*, pp. 688-689, Italy, July 2012.

[16] O. E. Krigolson, C. C. Williams, A. Norton, C. D. Hassall, and F. L. Colino, "Choosing MUSE: Validation of a low-cost, portable EEG system for ERP research," *Frontiers in Neuroscience*, vol. 11, p. 109, 2017.

[17] J. J. Bird, L. J. Manso, E. P. Ribiero, A. Ekart, and D. R. Faria, "A study on mental state classification using eeg-based brain-machine interface," in *Proceedings of the 9th International Conference on Intelligent Systems*, IEEE, 2018.

[18] C. Premebida, D. R. Faria, F. A. Souza, and U. Nunes, "Applying probabilistic Mixture Models to semantic place classification in mobile robotics," in *Proceedings of the IEEE/RSJ International Conference on Intelligent Robots and Systems, IROS 2015*, pp. 4265–4270, Germany, October 2015.

[19] C. Premebida, D. R. Faria, and U. Nunes, "Dynamic Bayesian network for semantic place classification in mobile robotics," *Autonomous Robots*, vol. 41, no. 5, pp. 1161–1172, 2017.

[20] T. Y. Chai, S. S. Woo, M. Rizon, and C. S. Tan, "Classification of human emotions from eeg signals using statistical features and neural network," *International Journal of Integrated Engineering (Penerbit UTHM)*, vol. 1, pp. 1–6, 2010.

[21] H. Tanaka, M. Hayashi, and T. Hori, "Statistical features of hypnagogic EEG measured by a new scoring system," *SLEEP*, vol. 19, no. 9, pp. 731–738, 1996.

[22] W. Zheng, J. Zhu, Y. Peng, and B. Lu, "EEG-based emotion classification using deep belief networks," in *Proceedings of the 2014 IEEE International Conference on Multimedia and Expo (ICME)*, pp. 1–6, Chengdu, China, July 2014.

[23] K. G. Jordan, "Emergency EEG and continuous EEG monitoring in acute ischemic stroke," *Journal of Clinical Neurophysiology*, vol. 21, no. 5, pp. 341–352, 2004.

[24] K. K. Ang, C. Guan, K. S. G. Chua et al., "Clinical study of neurorehabilitation in stroke using EEG-based motor imagery brain-computer interface with robotic feedback," in *Proceedings of the 2010 32nd Annual International Conference of the IEEE Engineering in Medicine and Biology Society, EMBC'10*, pp. 5549–5552, Argentina, September 2010.

[25] A. T. Tzallas, M. G. Tsipouras, and D. I. Fotiadis, "Epileptic seizure detection in EEGs using time-frequency analysis," *IEEE Transactions on Information Technology in Biomedicine*, vol. 13, no. 5, pp. 703–710, 2009.

[26] A. Aarabi, R. Grebe, and F. Wallois, "A multistage knowledge-based system for EEG seizure detection in newborn infants," *Clinical Neurophysiology*, vol. 118, no. 12, pp. 2781–2797, 2007.

[27] K.-M. Rytkönen, J. Zitting, and T. Porkka-Heiskanen, "Automated sleep scoring in rats and mice using the naive Bayes classifier," *Journal of Neuroscience Methods*, vol. 202, no. 1, pp. 60–64, 2011.

[28] L. Fraiwan, K. Lweesy, N. Khasawneh, H. Wenz, and H. Dickhaus, "Automated sleep stage identification system based on time-frequency analysis of a single EEG channel and random forest classifier," *Computer Methods & Programs in Biomedicine*, vol. 108, no. 1, pp. 10–19, 2012.

[29] Y. Peng and B.-L. Lu, "Immune clonal algorithm based feature selection for epileptic EEG signal classification," in *Proceedings of the 2012 11th International Conference on Information Science, Signal Processing and their Applications, ISSPA 2012*, pp. 848–853, Canada, July 2012.

[30] D. R. Faria, M. Vieira, F. C. C. Faria, and C. Premebida, "Affective facial expressions recognition for human-robot interaction," in *Proceedings of the 26th IEEE International Symposium on Robot and Human Interactive Communication, RO-MAN 2017*, pp. 805–810, Portugal, September 2017.

[31] J. J. Bird, A. Ekart, and D. R. Faria, "High resolution sentiment analysis by ensemble classification," in *Proceedings of the SAI Computing Conference, SAI, 2019*, 2019.

[32] C. Coppola, D. R. Faria, U. Nunes, and N. Bellotto, "Social activity recognition based on probabilistic merging of skeleton

features with proximity priors from RGB-D data," in *Proceedings of the 2016 IEEE/RSJ International Conference on Intelligent Robots and Systems, IROS 2016*, pp. 5055–5061, Republic of Korea, October 2016.

[33] D. A. Adama, A. Lotfi, C. Langensiepen, and K. Lee, "Human activities transfer learning for assistive robotics," in *Proceedings of the UK Workshop on Computational Intelligence*, pp. 253–264, Springer, 2017.

[34] S. W. Yahaya, C. Langensiepen, and A. Lotfi, "Anomaly detection in activities of daily living using one-class support vector machine," in *Proceedings of the UK Workshop on Computational Intelligence*, pp. 362–371, Springer, 2018.

[35] D. Ortega-Anderez, A. Lotfi, C. Langensiepen, and K. Appiah, "A multi-level refinement approach towards the classification of quotidian activities using accelerometer data," *Journal of Ambient Intelligence and Humanized Computing*, pp. 1–12, 2018.

[36] J. J. Bird, A. Ekárt, and D. R. Faria, "Learning from interaction: An intelligent networked-based human-bot and bot-bot chatbot system," in *Proceedings of the UK Workshop on Computational Intelligence*, pp. 179–190, Springer, 2018.

[37] P. A. Vikhar, "Evolutionary algorithms: a critical review and its future prospects," in *Proceedings of the 2016 International Conference on Global Trends in Signal Processing, Information Computing and Communication, ICGTSPICC 2016*, pp. 261–265, India, December 2016.

[38] C. Darwin, *On the origin of species, 1859*, Routledge, 2004.

[39] J. J. Bird, A. Ekart, and D. R. Faria, "Evolutionary optimisation of fully connected artificial neural network topology," in *Proceedings of the SAI Computing Conference, SAI, 2019*, 2019.

[40] D. H. Wolpert and W. G. Macready, "No free lunch theorems for optimization," *IEEE Transactions on Evolutionary Computation*, vol. 1, no. 1, pp. 67–82, 1997.

[41] D. E. Knuth, "Postscript about NP-hard problems," *ACM SIGACT News*, vol. 6, no. 2, pp. 15-16, 1974.

[42] Y. Bengio, I. J. Goodfellow, and A. Courville, "Deep learning," *Nature*, vol. 521, no. 7553, pp. 436–444, 2015.

[43] S. Kullback and R. A. Leibler, "On information and sufficiency," *Annals of Mathematical Statistics*, vol. 22, no. 1, pp. 79–86, 1951.

[44] F. Assunção, N. Lourenço, P. Machado, and B. Ribeiro, "DENSER: deep evolutionary network structured representation," *Genetic Programming and Evolvable Machines*, pp. 1–31, 2018.

[45] A. Martín, R. Lara-Cabrera, F. Fuentes-Hurtado, V. Naranjo, and D. Camacho, "EvoDeep: a new evolutionary approach for automatic deep neural networks parametrisation," *Journal of Parallel and Distributed Computing*, vol. 117, pp. 180–191, 2018.

[46] Y. Peng and B.-L. Lu, "Discriminative extreme learning machine with supervised sparsity preserving for image classification," *Neurocomputing*, vol. 261, pp. 242–252, 2017.

[47] A. Graves, N. Jaitly, and A.-R. Mohamed, "Hybrid speech recognition with deep bidirectional LSTM," in *Proceedings of the 2013 IEEE Workshop on Automatic Speech Recognition and Understanding, ASRU 2013*, pp. 273–278, Czech Republic, December 2013.

[48] P. R. Davidson, R. D. Jones, and M. T. R. Peiris, "Detecting behavioral microsleeps using EEG and LSTM recurrent neural networks," in *Proceedings of the 2005 27th Annual International Conference of the Engineering in Medicine and Biology Society, IEEE-EMBS 2005*, pp. 5754–5757, China, September 2005.

[49] Y. Freund and R. E. Schapire, "A decision-theoretic generalization of on-line learning and an application to boosting," *Journal of Computer and System Sciences*, vol. 55, no. 1, part 2, pp. 119–139, 1997.

[50] R. Rojas, "Adaboost and the super bowl of classifiers a tutorial introduction to adaptive boosting," Tech. Rep., Freie University, Berlin, Germany, 2009.

[51] T. Y. M. Chiu, T. Leonard, and K.-W. Tsui, "The matrix-logarithmic covariance model," *Journal of the American Statistical Association*, vol. 91, no. 433, pp. 198–210, 1996.

[52] C. van Loan, *Computational Frameworks for the Fast Fourier Transform*, vol. 10 of *Frontiers in Applied Mathematics*, Society for Industrial and Applied Mathematics (SIAM), Philadelphia, Pa, USA, 1992.

[53] T. Back, *Evolutionary Algorithms in Theory and Practice: Evolution Strategies, Evolutionary Programming, Genetic Algorithms*, Oxford University Press, 1996.

[54] M. Hussain, J. J. Bird, and D. R. Faria, "A study on cnn transfer learning for image classification," in *Proceedings of the UK Workshop on Computational Intelligence*, pp. 191–202, Springer, 2018.

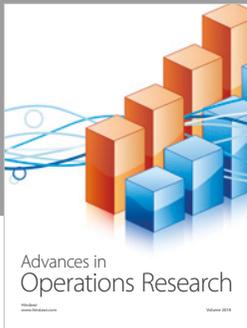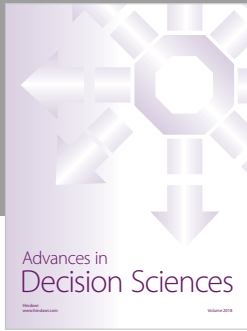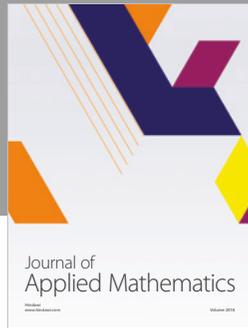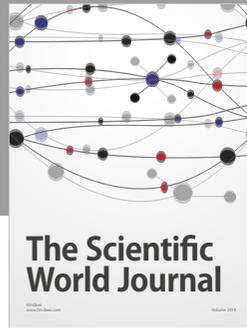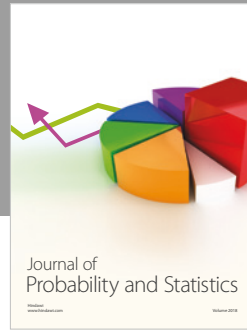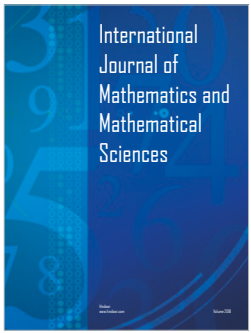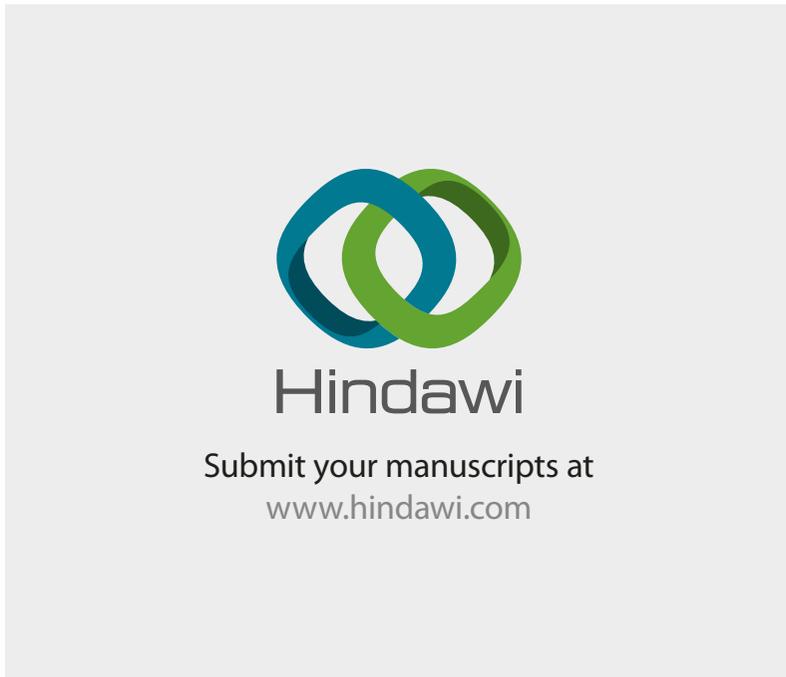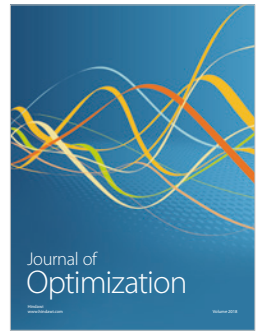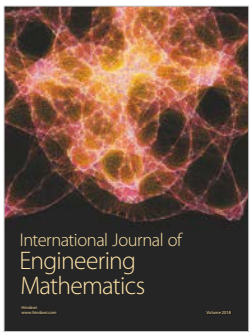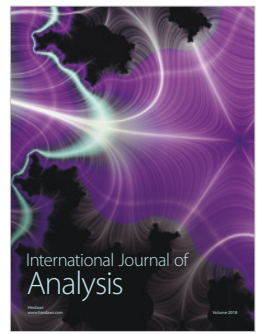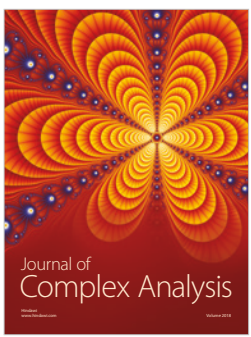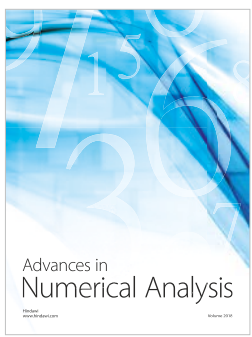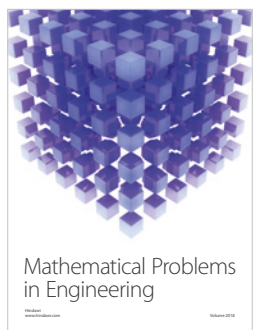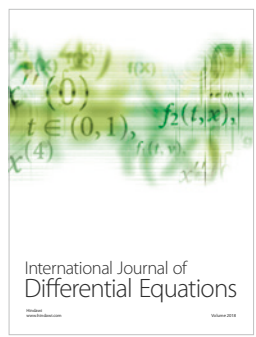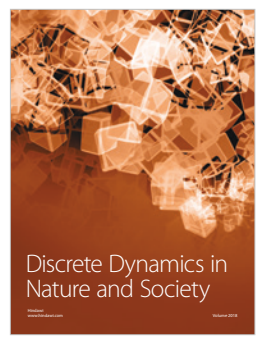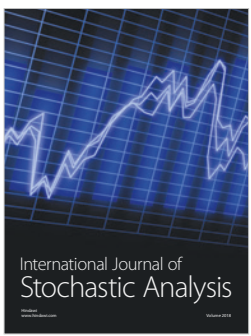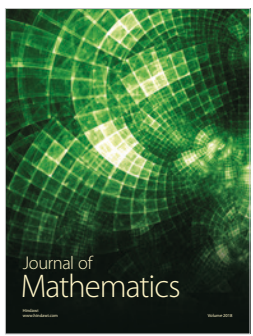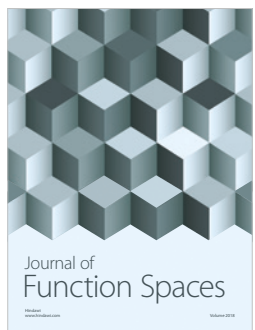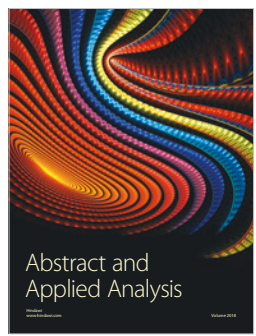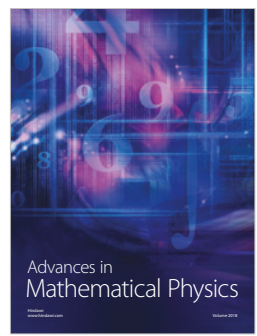